\documentclass{article}

\usepackage[preprint]{neurips_2021}

\usepackage[utf8]{inputenc} % allow utf-8 input
\usepackage[T1]{fontenc}    % use 8-bit T1 fonts
\usepackage{hyperref}       % hyperlinks
\usepackage{url}            % simple URL typesetting
\usepackage{booktabs}       % professional-quality tables
\usepackage{amsfonts}       % blackboard math symbols
\usepackage{nicefrac}       % compact symbols for 1/2, etc.
\usepackage{microtype}      % microtypography
\usepackage{xcolor}         % colors
\usepackage{graphicx}

\title{Architecture Matters: Investigating the Influence of Differential Privacy on Neural Network Design}

\author{%
  Felix Morsbach\\
  Institute AIFB\\
  Karlsruhe Institute of Technology\\
  Karlsruhe, Germany \\
  \texttt{felix.morsbach@kit.edu} \\
  \And
  Tobias Dehling\\
  Institute AIFB, KASTEL Security Research Labs\\
  Karlsruhe Institute of Technology\\
  Karlsruhe, Germany \\
  \texttt{dehling@kit.edu} \\
  \And
  Ali Sunyaev \\
  Institute AIFB, KASTEL Security Research Labs\\
  Karlsruhe Institute of Technology\\
  Karlsruhe, Germany \\
  \texttt{sunyaev@kit.edu} \\
}
\begin{document}

\maketitle

\begin{abstract}
One barrier to more widespread adoption of differentially private neural networks is the entailed accuracy loss. To address this issue, the relationship between neural network architectures and model accuracy under differential privacy constraints needs to be better understood. As a first step, we test whether extant knowledge on architecture design also holds in the differentially private setting. Our findings show that it does not; architectures that perform well without differential privacy, do not necessarily do so with differential privacy. Consequently, extant knowledge on neural network architecture design cannot be seamlessly translated into the differential privacy context. Future research is required to better understand the relationship between neural network architectures and model accuracy to enable better architecture design choices under differential privacy constraints. \end{abstract}

\section{Introduction}
Differential privacy has become the de facto standard for achieving data confidentiality in machine learning settings. 
The most prominent way to train differentially private neural networks is by clipping gradients in order to limit the impact of each data point and by adding noise to the gradient updates \cite{abadi_deep_2016}. 
While gradient clipping is also used in non-private training to avoid overfitting \cite{zhang_why_2020}, adding noise inherently reduces the accuracy of the machine learning model \cite{jayaraman_evaluating_2019}. 
This is considered as a trade-off between utility and differential privacy guarantees and is usually measured as accuracy loss, describing the difference between the accuracy with and without differential privacy constraints. 
Even on comparably simple tasks, such as the CIFAR-10 classification task, the accuracy loss can be significant.
This hinders the adoption of differentially private neural networks.

It is well known, that the architecture of a neural network can have a significant influence on the accuracy of the model. For example, expanding the size of a network in depth rather than width is usually said to increase accuracy \cite{goodfellow_multidigit_2014}, but also novel architectural features such as residual or dense connections can vastly improve model accuracy \cite{he_deep_2016, huang_densely_2017}. 
Yet, early works on differentially private neural networks do not seem to account for this in great detail. They either do not include different architectures in their benchmarks \cite{jayaraman_evaluating_2019} or find that the effect is not significant \cite{abadi_deep_2016}. 
There is, however, also contrary evidence. For instance, expanding the overall size of the network in a differentially private setting exhibits an inflection point \cite{papernot_making_2019}. Increasing the network size beyond this inflection point reduces model accuracy, which is not the case in the plain setting.
Furthermore, the choice of an activation function plays an important role. Bounded activation functions (e.~g.,~ tanh) consistently outperform unbounded ones (e.~g.,~ ReLU) due to the phenomena of exploding activations during differentially private training \cite{papernot_tempered_2021}.

Nevertheless, even with these findings taken into account, there still remains a significant gap between the accuracy of differentially private and non-private neural networks, even on simple tasks. 
Extant research has already shown that the choice of an appropriate activation function differs between the non-private and differentially private setting and that choosing an activation function appropriate for differentially private neural networks can reduce the incurred accuracy loss \cite{papernot_tempered_2021}. 
However, there are a multitude of other architectural features, such as the number, ordering, type and configuration of layers, whose impact on model accuracy under differential privacy constraints has not been really understood.
Consequently, we investigate whether architectural features other than the activation function of a neural network affect the accuracy loss incurred by differential privacy constraints.

If we can show that more architectural features than the activation function impact model accuracy, this would imply that the accuracy of differentially private models can be improved by carefully tuning a neural network architecture for differential privacy instead of simply copying architectures and hyperparameters that work well without differential privacy constraints \cite{vanderveen_three_2018}.

We designed and carried out an experiment to test whether the architecture affects accuracy loss under differential privacy constraints.
Our findings show, not only, that the network architecture has an influence on the accuracy loss incurred by differential privacy constraints, but also, that the suitability of architecture choices is sensitive to variations in the targeted level of differential privacy.
We elaborate the implications of our findings and argue what future research is necessary to reduce the accuracy loss incurred by differential privacy constraints in order to improve the applicability of privacy-preserving machine learning.

\section{Background}
Differentially private stochastic gradient descent (DP-SGD) is an algorithm for training neural networks with $(\epsilon, \delta)$-differential privacy guarantees \cite{abadi_deep_2016, dwork_algorithmic_2014}. 
DP-SGD extends classic stochastic gradient descent in two ways. 
First, gradients are clipped on a per-example basis to a $l_2$ norm, which is set through a clipping threshold $C$. 
Second, random noise is added to the gradient update calibrated via the standard deviation $\sigma$, also called noise multiplier. 
Thus, DP-SGD adds two additional hyperparameters to the training algorithm. 

The privacy level $(\epsilon, \delta)$ of a model trained with DP-SGD is dependent on the batch size, the noise multiplier, the number of epochs, and the number of training examples. 
Hence, the privacy level for a given set of these hyperparameters can be calculated upfront without the need to actually train the model. 
Since the privacy level is independent of the architecture of a given model,  different architectures can be compared at a fixed privacy level. 

\section{Experiments}
In order to answer our research question, we need to show that the architecture of a neural network does have an effect on the model’s accuracy loss under differential privacy constraints. We hypothesize that given a set of neural architectures for a given machine learning task, the architecture which performs best without differential privacy constraints does not necessarily also performs best under differential privacy constraints. Or formulated differently, let there be two neural network architectures $A_1$ and $A_2$ and let $U(A)$ denote the accuracy of architecture $A$ without differential privacy and $U_d(A)$ the accuracy with differential privacy. We assume the case exists that $U(A_1) > U(A_2)$ and $U_d(A_1) < U_d(A_2)$. 

To test our hypothesis, we conducted an experiment on the standard CIFAR-10 image classification task which contains $60000$ color images in $10$ classes \cite{krizhevsky_learning_2009}. We used Tensorflow \cite{abadi_tensorflow_2016} and the Tensorflow-Privacy extension \cite{andrew_tensorflow_2018} as machine learning libraries for the implementation. For our experiments we chose $8$ convolutional neural network architectures, including prominent examples from the literature such as the LeNet-5 \cite{lecun_gradientbased_1998}, but also architectures from other well-cited works \cite{arachchige_local_2019, mcmahan_communicationefficient_2017, phan_adaptive_2017} and own creations. The architectures differ mostly in the size and number of convolution, pooling, and fully-connected layers. We chose convolutional architectures as an initial setting, as their architecture configuration space is more complex and interesting compared to simple feedforward networks with only fully connected layers. 
See Table \ref{tab:architectures} for a full definition of the architectures used. We trained each model architecture for $100$ epochs with DP-SGD, with a batch size of $250$, $5$ micro batches, a fixed learning rate of $0.1$, a clipping threshold of $1.0$ and at two noise multipliers of $0.01$ and $0.1$. 

\begin{table}
  \caption{The architectures used for the experiment}
  \label{tab:architectures}
  \centering
  \begin{tabular}{llll}
    \toprule
    \multicolumn{2}{c}{$\#1$} & \multicolumn{2}{c}{$\#2$} \\
    \cmidrule(lr\cmidrulekern){1-2} \cmidrule(lr\cmidrulekern){3-4}
    Layer type &    Parameters &    Layer type &    Parameters \\
    \cmidrule(lr\cmidrulekern){1-2} \cmidrule(lr\cmidrulekern){3-4}
    Convolution     & 32 filters of 3x3, ReLU   & Convolution       & 32 filters of 3x3, ReLU   \\
    Max-Pooling     & 2x2                       & Max-Pooling       & 2x2                       \\
    Convolution     & 64 filters of 3x3, ReLU   & Convolution       & 64 filters of 3x3, ReLU   \\
    Fully connected & 64 units, ReLU            & Max-Pooling       & 2x2                       \\
    Softmax         & 10 units                  & Convolution       & 64 filters of 3x3, ReLU   \\
                    &                           & Fully connected   & 64 units, ReLU            \\
                    &                           & Softmax           & 10 units                  \\
    \addlinespace[1em]
    \multicolumn{2}{c}{$\#3$} & \multicolumn{2}{c}{$\#4$, \cite{lecun_gradientbased_1998} but with ReLU} \\
    \cmidrule(lr\cmidrulekern){1-2} \cmidrule(lr\cmidrulekern){3-4}
    Layer type &    Parameters &    Layer type &    Parameters \\
    \cmidrule(lr\cmidrulekern){1-2} \cmidrule(lr\cmidrulekern){3-4}
    Convolution         & 32 filters of 3x3, ReLU   & Convolution       & 6 filters of 5x5, ReLU    \\
    Max-Pooling         & 2x2                       & Avg-Pooling       & 2x2, stride 2             \\
    Convolution         & 64 filters of 3x3, ReLU   & Convolution       & 16 filters of 5x5, ReLU   \\
    Max-Pooling         & 2x2                       & Avg-Pooling       & 2x2, stride 2             \\
    Convolution         & 64 filters of 3x3, ReLU   & Fully connected   & 120 units, ReLU           \\
    Fully connected     & 128 units, ReLU           & Fully connected   & 84 units, ReLU            \\
    Softmax             & 10 units                  & Softmax           & 10 units                  \\
    \addlinespace[1em]
    \multicolumn{2}{c}{$\#5$} & \multicolumn{2}{c}{$\#6$, \cite{arachchige_local_2019}} \\
    \cmidrule(lr\cmidrulekern){1-2} \cmidrule(lr\cmidrulekern){3-4}
    Layer type &    Parameters &    Layer type &    Parameters \\
    \cmidrule(lr\cmidrulekern){1-2} \cmidrule(lr\cmidrulekern){3-4}
    Convolution         & 32 filters of 5x5, ReLU   & Convolution       & 32 filters of 3x3, ReLU   \\
    Avg-Pooling         & 2x2, stride 2             & Convolution       & 32 filters of 3x3, ReLU   \\
    Convolution         & 64 filters of 5x5, ReLU   & Max-Pooling       & 2x2                       \\
    Avg-Pooling         & 2x2, stride 2             & Dropout           & $0.25$                    \\
    Fully connected     & 200 units, ReLU           & Convolution       & 64 filters of 3x3, ReLU   \\
    Fully connected     & 100 units, ReLU           & Convolution       & 64 filters of 3x3, ReLU   \\
    Softmax             & 10 units                  & Max-Pooling       & 2x2                       \\
                        &                           & Fully connected   & 512 units, ReLU           \\
                        &                           & Dropout           & $0.5$                     \\
                        &                           & Softmax           & 10 units                  \\
    \addlinespace[1em]
    \multicolumn{2}{c}{$\#7$, \cite{mcmahan_communicationefficient_2017}} & \multicolumn{2}{c}{$\#8$, \cite{phan_adaptive_2017}} \\
    \cmidrule(lr\cmidrulekern){1-2} \cmidrule(lr\cmidrulekern){3-4}
    Layer type &    Parameters &    Layer type &    Parameters \\
    \cmidrule(lr\cmidrulekern){1-2} \cmidrule(lr\cmidrulekern){3-4}
    Convolution         & 32 filters of 5x5, ReLU   & Convolution       & 32 filters of 5x5, ReLU   \\
    Avg-Pooling         & 2x2, stride 2             & Convolution       & 64 filters of 5x5, ReLU   \\
    Convolution         & 64 filters of 5x5, ReLU   & Fully connected   & 384 units, ReLU           \\
    Avg-Pooling         & 2x2, stride 2             & Fully connected   & 192 units, ReLU           \\
    Fully connected     & 512 units, ReLU           & Softmax           & 10 units                  \\
    Softmax             & 10 units                  &                   &                           \\
    \addlinespace[1em]
    \bottomrule
  \end{tabular}
\end{table}

\section{Results}
\begin{figure}
\center
\includegraphics[width=\linewidth]{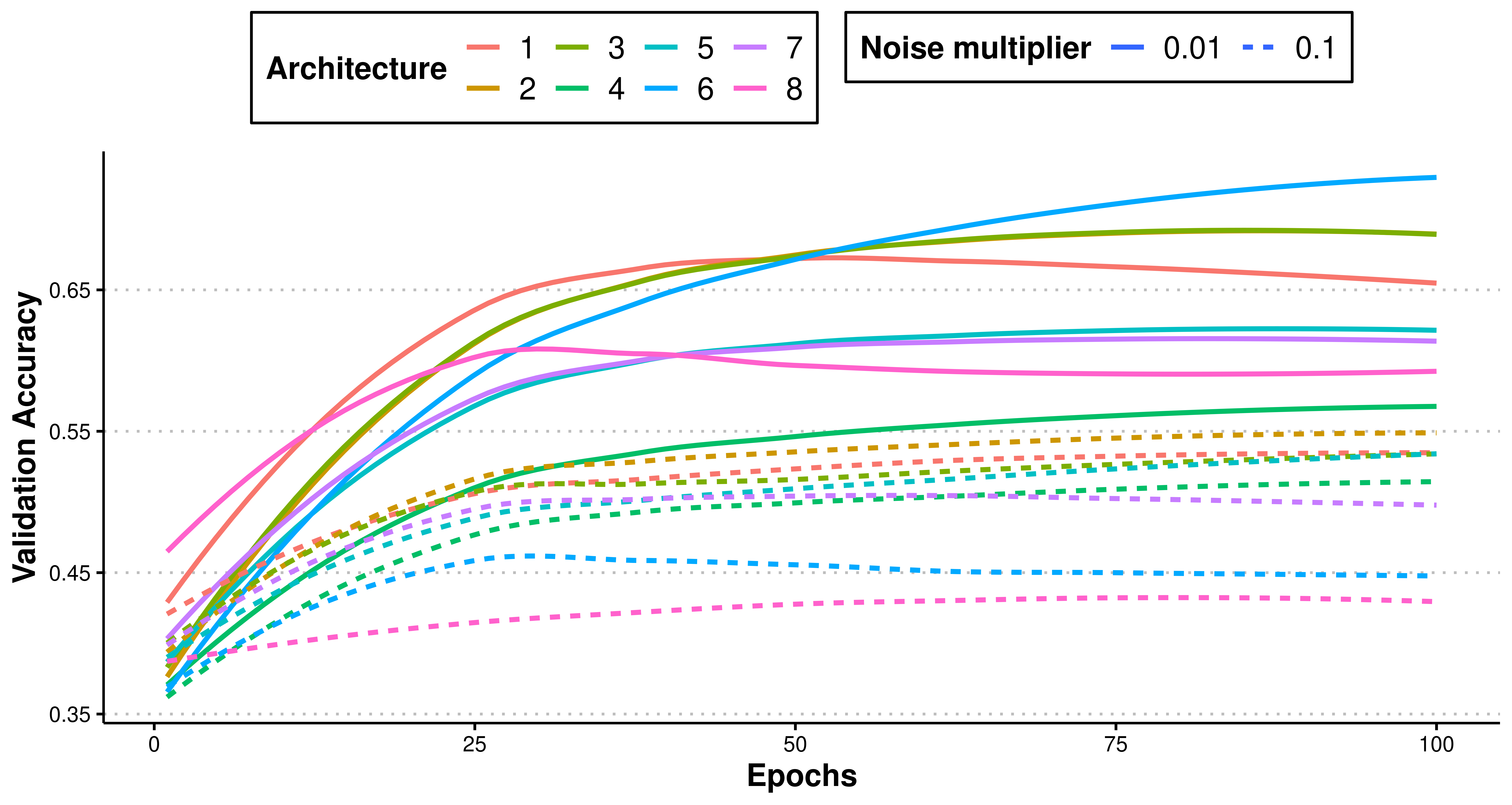}
\caption{Results of training 8 neural architectures for 100 epochs on the CIFAR-10 image classification task with DP-SGD and a batch size of $250$, $5$ micro batches, a fixed learning rate of $0.1$ and a clipping threshold of $1.0$, at two different noise multipliers ($0.01$ and $0.1$). (\textit{Best viewed in color.})}
\label{pic:results2}
\end{figure}

Figure \ref{pic:results2} shows the training results of the $8$ different architectures on the CIFAR-10 machine learning task, repeated three times in identical settings with two different noise multipliers. We can clearly see that with a low noise multiplier model $6$ performs best, but with a high noise multiplier model $6$ comes in second to last while model $2$ performs best. As the noise multiplier is one of the hyperparameters that determines the privacy level (besides epochs, batch size, and training set size, which are identical for all results), this means that at different privacy levels, different architectures perform best.
In turn, this shows that, in a given set of architectures, it is not always the case that a single architecture will perform best for all differential privacy settings. Therefore, it is also not guaranteed that an architecture that performs best without differential privacy necessarily performs best in the differential privacy setting since its performance will be dependent on the targeted privacy level.
The raw results and the code used to generate the graph can be found on GitHub \footnote{\tiny{\url{https://github.com/FMorsbach/ArchitectureMatters}}}.

\section{Conclusion}
In this paper, we investigated whether the architecture of a neural network can influence the accuracy loss incurred through differential privacy constraints. 
Our findings show that this is the case; neural network architectures that perform well in the non-private setting will not necessarily perform well in the differentially private setting. 
Furthermore, we found that which architecture performs best also depends on the chosen level of differential privacy.

The implications of our findings are two-fold:
First, for research, our findings show that architectures are only comparable at the same privacy level, as the relative ranking of architectures might differ across privacy levels.
Second, for practitioners, our findings show that best practices, experiences, or architectures from the non-private setting cannot be easily transferred to the design of differentially private neural networks. Rather, the model architecture has to be designed and tuned for specific differential privacy settings; differential privacy cannot be treated as an afterthought.

Moreover, the design of neural networks is a challenging task that either takes a lot of resources to try many different architectures through neural architecture search \cite{elsken_neural_2019} or requires a good understanding of neural network design from the modeler.
Doing a neural architecture search in the plain setting requires additional computations, but an extensive search will not hurt the final accuracy of the model. 
Searching for an architecture or optimizing hyperparameters in the differential privacy setting does, however, consume privacy budget \cite{vanderveen_three_2018}. 
Therefore, spending more time on architecture search or hyperparameter optimization will decrease the privacy budget available for the actual training, which will decrease the number of epochs available for training; hence, it will probably reduce the accuracy of the final model. 
As a consequence, neural architecture search is not easily applicable in the differential privacy setting. 

Instead, a good understanding of how to design neural network architectures in the differential privacy setting is needed in order to maximize the available privacy budget for the actual training. 
But as shown by our findings, the experience from the design of neural network architectures without differential privacy constraints can only be transferred to the design of differentially private neural network architectures to a limited degree.
Therefore, we need to derive new best practices for the design of architectures tailored specifically for differentially private neural networks, in order to improve the applicability of privacy-preserving machine learning by reducing the accuracy loss incurred by differential privacy constraints. 

In our future work, we will derive best practices for the design of neural architectures under differential privacy constraints. We will set up additional experiments across multiple classic machine learning benchmarks. Subsequently, we will analyse the results, derive a set of candidate best practices and test those on different benchmarks. We aim to also incorporate qualitative data from expert interviews to even further advance the best practices for designing neural network architectures that perform well under differential privacy constraints.

\begin{ack}
This work was supported by funding from the topic Engineering Secure Systems of the Helmholtz Association (HGF) and by KASTEL Security Research Labs.
\end{ack}

\bibliography{neurips.priml.bib}
\bibliographystyle{plain}

\end{document}